\documentclass[letterpaper, 10 pt, journal, twoside]{IEEEtran}  % Comment this line out if you need a4paper

\IEEEoverridecommandlockouts                              % This command is only needed if
\usepackage{graphics} % for pdf, bitmapped graphics files
\usepackage{epsfig} % for postscript graphics files
\usepackage{mathptmx} % assumes new font selection scheme installed
\usepackage{times} % assumes new font selection scheme installed
\usepackage{amsmath} % assumes amsmath package installed
\usepackage{amssymb}  % assumes amsmath package installed
\usepackage{subfig}
\usepackage{cite}
\usepackage[amssymb]{SIunits}
\usepackage{graphicx}
\usepackage{amssymb}
\usepackage{multirow}
\usepackage{wrapfig}
\usepackage{seqsplit}

\usepackage{amsmath}
\usepackage{bm} % optional
\usepackage{footnote}
\usepackage{threeparttable}
\usepackage{epsfig}
\usepackage{graphicx}
\usepackage{amsmath}
\usepackage{balance}

\usepackage{url}            % simple URL typesetting
\usepackage{booktabs}       % professional-quality tables
\usepackage{amsfonts}       % blackboard math symbols
\usepackage{nicefrac}       % compact symbols for 1/2, etc.
\usepackage{microtype}      % microtypography
\usepackage{lipsum}
\usepackage{paralist}
\usepackage{color}

\makeatletter
\let\NAT@parse\undefined
\makeatother
\usepackage[colorlinks = true,
            linkcolor = blue,
            urlcolor  = blue,
            citecolor = blue,
            anchorcolor = blue]{hyperref}

\usepackage{array}
\newcommand{\PreserveBackslash}[1]{\let\temp=\\#1\let\\=\temp}
\newcolumntype{C}[1]{>{\PreserveBackslash\centering}p{#1}}
\newcolumntype{R}[1]{>{\PreserveBackslash\raggedleft}p{#1}}
\newcolumntype{L}[1]{>{\PreserveBackslash\raggedright}p{#1}}

\newcommand{\rom}[1]{\uppercase\expandafter{\romannumeral #1\relax}}

\newcommand{\fref}[1]{Fig.~\ref{#1}}
\newcommand{\sref}[1]{Section~\ref{#1}}
\newcommand{\tref}[1]{Table~\ref{#1}}
\newcommand{\fix}[1]{#1}

\newcommand{\etal}{\textit{et al.}~}
\newcommand{\ie}{{i.e.},~}

\graphicspath{%
    {./figures/}%
}

\title{\LARGE \bf AirCode: A Robust Object Encoding Method}

\author{Kuan Xu$^{1}$, Chen Wang$^{2}$, Chao Chen$^{1}$, Wei Wu$^{1}$, and Sebastian Scherer$^{2}$% <-this % stops a space
\thanks{Manuscript received September 9, 2021; accepted December 30, 2021. This letter was recommended for publication by Associate Editor I. Gilitschenski and Editor C. Cadena Lerma upon evaluation of the reviewers' comments. Source code and pre-trained models are available at \url{https://github.com/wang-chen/AirCode}. Video is available at \url{https://youtu.be/ZhW4Qk1tLNQ}. (Corresponding author: Chen Wang.)}%Use only for final RAL version
\thanks{$^{1}$Kuan Xu, Chao Chen, and Wei Wu are with the Robot R\&D Department, Geekplus Corp., Beijing 100107, China (email: xukuanhit@gmail.com; chenchao@geekplus.com; merlinwu@geekplus.com).}%
\thanks{$^{2}$Chen Wang and Sebastian Scherer are with the Robotics Institute, Carnegie Mellon University, Pittsburgh, PA 15213, USA (email: chenwang@dr.com; basti@andrew.cmu.edu).}%
\thanks{Digital Object Identifier 10.1109/LRA.2022.3141221}
}

\begin{document}

\maketitle
% \thispagestyle{empty}
% \pagestyle{empty}

%%%%%%%%%%%%%%%%%%%%%%%%%%%%%%%%%%%%%%%%%%%%%%%%%%%%%%%%%%%%%%%%%%%%%%%%%%%%%%%%
\begin{abstract}
Object encoding and identification are crucial for many robotic tasks such as autonomous exploration and semantic relocalization. Existing works heavily rely on the tracking of detected objects but have difficulty recalling revisited objects precisely. In this paper, we propose a novel object encoding method, which is named as AirCode, based on a graph of key-points. To be robust to the number of key-points detected, we propose a feature sparse encoding and object dense encoding method to ensure that each key-point can only affect a small part of the object descriptors, leading it to be robust to viewpoint changes, scaling, occlusion, and even object deformation. In the experiments, we show that it achieves superior performance for object identification than the state-of-the-art algorithms and is able to provide reliable semantic relocalization. It is a plug-and-play module and we expect that it will play an important role in various applications.
\end{abstract}

\begin{IEEEkeywords}
Visual Learning, Recognition, Object Encoding
\end{IEEEkeywords}

%%%%%%%%%%%%%%%%%%%%%%%%%%%%%%%%%%%%%%%%%%%%%%%%%%%%%%%%%%%%%%%%%%%%%%%%%%%%%%
\section{Introduction}

\IEEEPARstart{O}{bject} encoding and identification are of great importance to many robotic tasks such as autonomous exploration and semantic re-localization in simultaneous localization and mapping (SLAM).
For example, in autonomous exploration, an efficient and robust object encoding benefits the decision process when a robot revisits a specific landmark object \cite{wang2020visual,wang2021unsupervised}.
Without the capability of \fix{place recognition}, a semantic SLAM system may easily drift and subsequently lead to an unreliable localization \cite{wang2020online}.
However, existing object encoding methods easily produce false matches due to viewpoint or scaling changes, hence a robust and efficient object encoding method is necessary for many robotic applications.

To be efficient, object matching in SLAM is often based on key-point features \cite{dubuisson1994modified}, as the feature-based SLAM methods \cite{mur2017orb} are still widely used.
Inspired by the recent progress in deep learning-based key-point detector \cite{detone2018superpoint} and feature matching \cite{sarlin2020superglue}, it becomes intuitive to encode an object via a group of key-points in an end-to-end manner, where the key-points on the same object form a graph neural network.
Therefore, we can take the graph embeddings as the object descriptors.

However, this is not straightforward and very difficult, since the number of detected object key-points are affected by many factors such as illumination and object deformation.
Moreover, during robot exploration, robots often observe part of the objects due to occlusion and different viewpoints, resulting in that the object key-points only have a small overlap across different frames.
Therefore, the key-points graph embedding will be easily affected, which makes it difficult to directly apply a graph network \cite{wang2020lifelong}.
To solve this problem, we argue that \fix{a single key-point should nearly have no effect on the object embedding}.
This means that only a few positions of an object descriptor can be affected if a key-point is added into or removed from an object.
To achieve this, we propose a sparse object encoding method, which is robust to the change of viewpoint and object deformation.

\begin{figure}[t]
    \centering
    \includegraphics[width=1\linewidth]{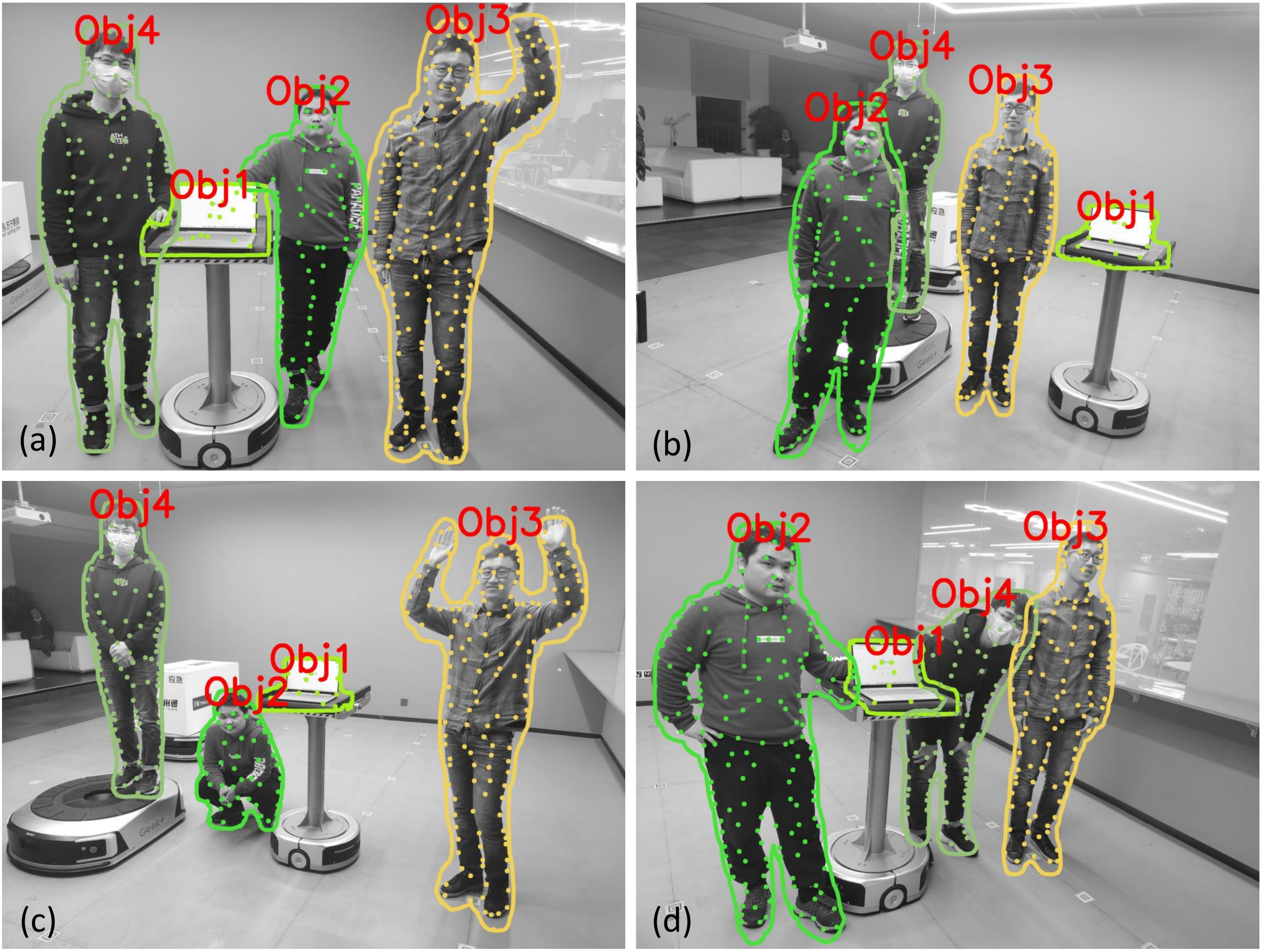}
    \caption{An object matching example. Four objects including 1 rigid object (laptop) and 3 non-rigid objects (human) are identified. Our encoding method is insensitive to viewpoint (Obj 1 in (b) and (d)), scaling (Obj 3 and Obj 4), occlusion (Obj 4), and even posture change (Obj 2 and Obj 3).}
    \label{fig:demo}
\end{figure}

In summary, the main contributions of this paper are
\begin{itemize}
    \item We introduce a simple yet effective pipeline, AirCode, for robust object encoding and present a plug-and-play sparsity layer to cluster similar key-points.
    \item We propose a sparse loss for key-point features and a dense loss for object descriptors to ensure the object encoder to be insensitive to specific feature points.
    \item AirCode provides reliable object matching as shown in \fref{fig:demo} and we demonstrate its effectiveness in semantic relocalization compared to the state-of-the-art methods.
\end{itemize}

\section{Related Work}

We will review both object encoding and visual place recognition methods, since some place recognition algorithms can also be applied to object matching.
General detection networks such as Mask R-CNN \cite{he2017mask} will not be included, since they are difficult to be used for object matching.

\subsection{Handcrafted Feature-based Methods}

Fast appearance-based mapping (FABMAP) \cite{glover2012openfabmap} is one of the most classic approaches on handcrafted features. The authors train a visual vocabulary of SURF feature \cite{bay2006surf} using hierarchical k-means clustering to discretize the high dimensional representation, and then the offline trained visual vocabulary can be used to match features to identify revisited objects. Furthermore, DBoW2 \cite{galvez2012bags} follows this idea and employs ORB binary descriptor \cite{rublee2011orb} to achieve a faster speed.  In \cite{garcia2018ibow}, a dynamic island algorithm is proposed to maintain the vocabulary by grouping similar images, in which repetitive registration of the same descriptor across multiple frames is able to be identified. The strategy of removing redundant descriptors can keep the database at a small scale.

Schlegel \etal \cite{schlegel2018hbst} introduce an online trained Hamming distance embedded binary search tree (HBST) for image retrieval, which is much faster than traditional FLANN matching methods. However, due to the use of raw descriptors to build the incremental visual vocabulary tree, its memory cost is huge. Reducing the dimension of local features is also an alternative solution. Inspired by the image encoding methods, Carrasco \etal  \cite{carrasco2016global} apply hash coding to the local features array, which makes the features extracted from a single image more compact. Gehrig \etal  \cite{gehrig2017visual} reduce the dimension of BRISK \cite{leutenegger2011brisk} feature with principal component analysis (PCA). They directly employ k-nearest neighbor (K-NN) search on projected descriptors, which makes the query speed several milliseconds faster than searching on pre-trained visual vocabulary. However, those methods are based on handcrafted feature extraction, which is not robust enough when the environment is changed \cite{sarlin2020superglue}.

Tsintotas \etal \cite{tsintotas2018assigning} adopt a dynamic quantization strategy to train the feature online, in which a growing neural gas \cite{fritzke1995growing} network is applied.
To utilize both structural and visual information, Stumm \textit{et al.} introduce covisibility graph to represent visual observations \cite{stumm2016robust}.
To jointly consider external information, Schlegel \textit{et al.} embed continuous and selector cues by augmenting each feature descriptor with a binary string \cite{schlegel2019adding}.
The feature-based methods perform not well when the descriptors are not discriminative enough \cite{geiger2012we}, hence we argue that an object-level place recognition is necessary.

\subsection{Deep Feature-based Methods}

Convolutional neural network (CNN) has made great progress in many computer vision tasks and this triggers another research trend \cite{wang2019kervolutional}. For example, Z. Chen \etal \cite{chen2017deep} introduce a multi-scale feature encoding method, in which two CNN architectures are trained and generate features. The viewpoint invariant features provide a huge performance boost. Relja et al. propose NetVLAD \cite{arandjelovic2016netvlad}, in which the authors train a convolutional neural network to replace the parameters of the vector of locally aggregated descriptors (VLAD). Instead of relying on a single method, Hausler \etal \cite{hausler2019multi} combine multiple methods of image processing for visual place recognition, including the sum of absolute difference, histogram of oriented gradients (HOG) \cite{dalal2005histograms}, spatial Max pooling, and spatial Arg-Max pooling.

To overcome the problem of training data deficiency and multi-modality input dissimilarity, an RGB-D object recognition framework is proposed to effectively embed depth and point cloud data into the RGB domain \cite{zaki2016convolutional}.
Zaki \etal incorporated depth information into a transferred CNN model from image recognition by rendering objects from a canonical perspective and colorizing the depth channel according to distance from the object center \cite{schwarz2015rgb}.
HP-CNN \cite{zaki2019viewpoint} presents a multi-scale object feature representation based on a multi-view 3D object pose using RGB-D sensors.

Our object encoding method is established on deep feature points and descriptors, which has received increasing attention recently.
For example, SuperPoint \cite{detone2018superpoint} proposes a self-supervised framework for training interest point detectors and descriptors. SuperGlue \cite{sarlin2020superglue} introduces a graph attention model for feature matching. Intuitively, the interest points and their descriptors form a large \fix{fully connected} temporally growing graph, in which the feature points are nodes, and their associated descriptors are the node features.

\begin{figure}[t]
    \centering
    \includegraphics[width=0.95\linewidth]{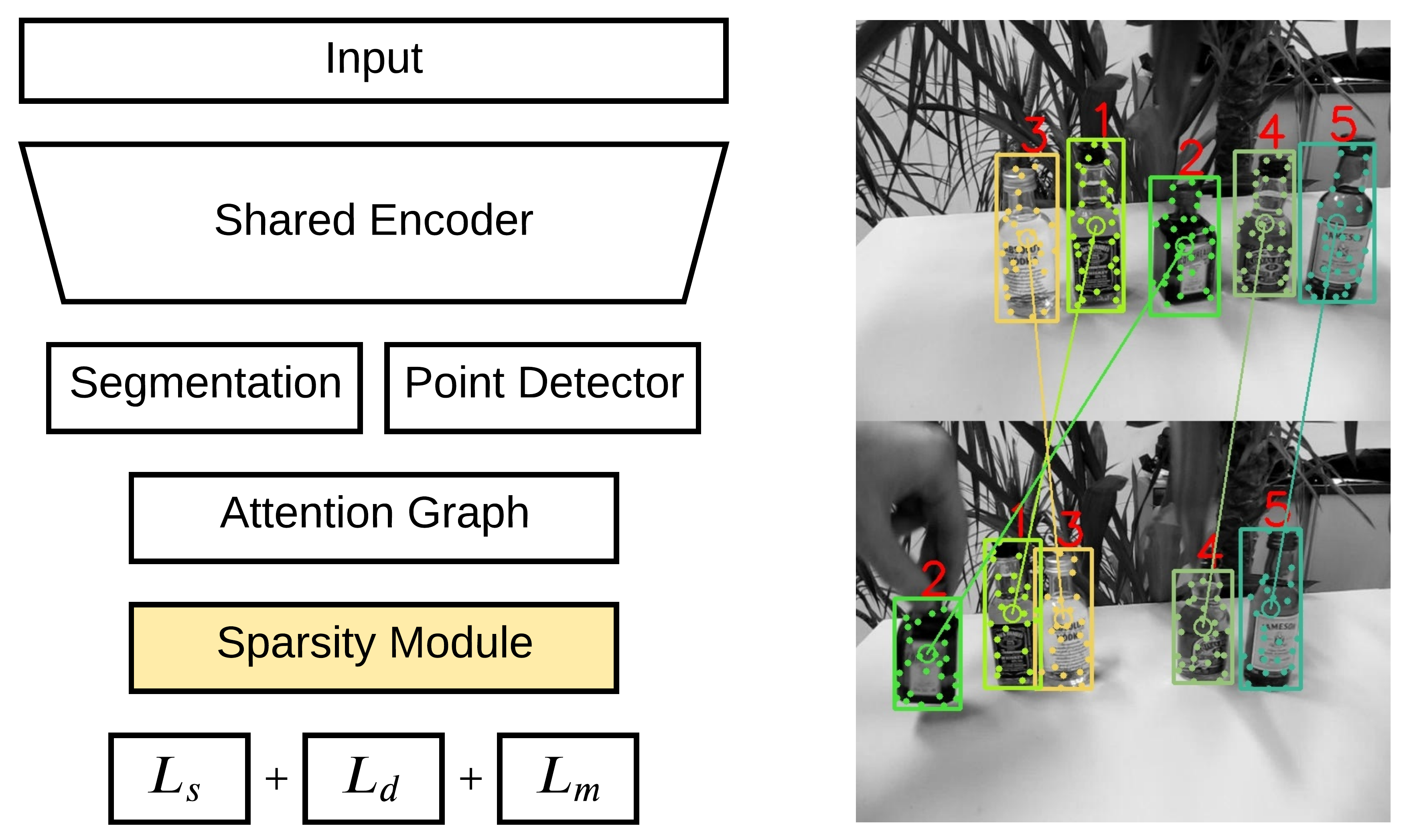}
    \caption{The network structure of AirCode and an example of object matching from the OTB dataset. The detailed structure of the sparsity module is presented in \fref{fig:sparsity-module}.}
    \label{fig:structure}
\end{figure}

\section{Methodology}

\fix{

\subsection{Network Structure}

As shown in \fref{fig:structure}, we propose a simple yet effective framework, AirCode, for object encoding.
For each image, we first segment/detect objects and get the feature points on each object.
To achieve this, we adopt two off-the-shelf headers with a shared backbone (VGG-16 \cite{simonyan2014very}), \ie a segmentation header (Mask R-CNN \cite{he2017mask}) for object detection/segmentation and a point detector (SuperPoint \cite{detone2018superpoint}) for points extraction.

We then take each object as a fully connected graph, where the nodes are the feature points on that object.
Assume that there are $M$ key-points on an object, where each point is denoted as its position $\mathbf{p}_i=(x,y)$ and associated descriptor $\mathbf{d}_i\in\mathbb{R}^{N_p}$ ($i\in[1,M]$), where $N_p$ is the dimension of the point descriptor.
Our goal is to generate an object descriptor $\mathbf{D}_k\in\mathbb{R}^{N_o}$ for robust object matching based on the object key-points, where $N_o$ ($N_o>N_p$) is the dimension of the object descriptor.
For simplicity, we skip the details of the segmentation header and point detector, which are pre-trained models.
Instead, we present the sparsity module and loss functions including sparse loss $L_s$, dense loss $L_d$, and matching loss $L_m$.
}

\fix{
\subsection{Graph Embedding with Sparse Node Features}

\subsubsection{Motivation}\label{sec:motivation}

Inspired by the fact that humans can easily identify an object by recognizing both local distinctive features and global structure \cite{tarr2017concurrent}, we believe that a robust object encoding method should have the following properties: (\rom{1}) An object descriptor should be able to encode both the local key-point features and the global structural information; (\rom{2}) The object descriptor should not be significantly affected when a key-point is added or removed, resulting in the insensitivity to viewpoint changes and object deformation.

Therefore, we argue that if only specific positions of the object descriptor can be affected by a key-point (in other words, a key-point has a sparse effect on its object descriptor), then we will be able to obtain a viewpoint invariant object encoding method.
This inspires us to formulate the object encoding as a problem of graph sparse embedding.

\subsubsection{Attention Graph}\label{sec:node-encoding}

To encode the local pattern and global structural information simultaneously, we first encode node features $\mathbf{x}_i$ for each point as the concatenation of the point positions $\mathbf{p}_i$ and descriptor $\mathbf{d}_i$, so that
\begin{equation}\label{eq:node-feature}
   \mathbf{x}_i = [\mathbf{d}_i || \operatorname{MLP}(\mathbf{p}_i)], \quad \mathbf{x}_i \in \mathbb{R}^{N_n},
\end{equation}
where $N_n = N_p + N_m$, $||$ is the concatenation operator, $\operatorname{MLP}: \mathbb{R}^{2}\mapsto\mathbb{R}^{N_m}$ is a multi-layer perceptron module. In practice, the position $\mathbf{p}_i$ is normalized to $[-1,1]$ by the object size, which takes the object center as the origin.

We assume that each object is a fully connected graph, where the edge connection weights are learned from an attention mechanism \cite{vaswani2017attention}. Specifically, the edge weight $\alpha_{ij}$ between node $\mathbf{x}_i$ and node $\mathbf{x}_j$ is given as:
\begin{subequations}\label{eq:graph_edg_q}
    \begin{align}\label{eq:graph_edg_k}
        \mathbf{\alpha}_{ij} &= \mathbf{q}_i^\mathsf{T} \cdot \mathbf{k}_j,\\
        \mathbf{q}_i &=\mathbf{W}_1 \cdot \mathbf{x}_i + \mathbf{b}_1,\\
        \mathbf{k}_j &=\mathbf{W}_2 \cdot \mathbf{x}_j + \mathbf{b}_2,
    \end{align}
\end{subequations}
where $\mathbf{W}_1$, $\mathbf{W}_2$, $\mathbf{b}_1$, and $\mathbf{b}_2$ are learnable parameters, which are adopted from the graph attention module in \cite{sarlin2020superglue}.
This encodes the object structure into the node embedding.
We refer to \cite{velickovic2018graph} for more details about the graph attention network.

}

\subsubsection{Sparsity Layer} \label{sec:sparsity}
\begin{figure}[t]
    \centering
    \includegraphics[width=0.6\linewidth]{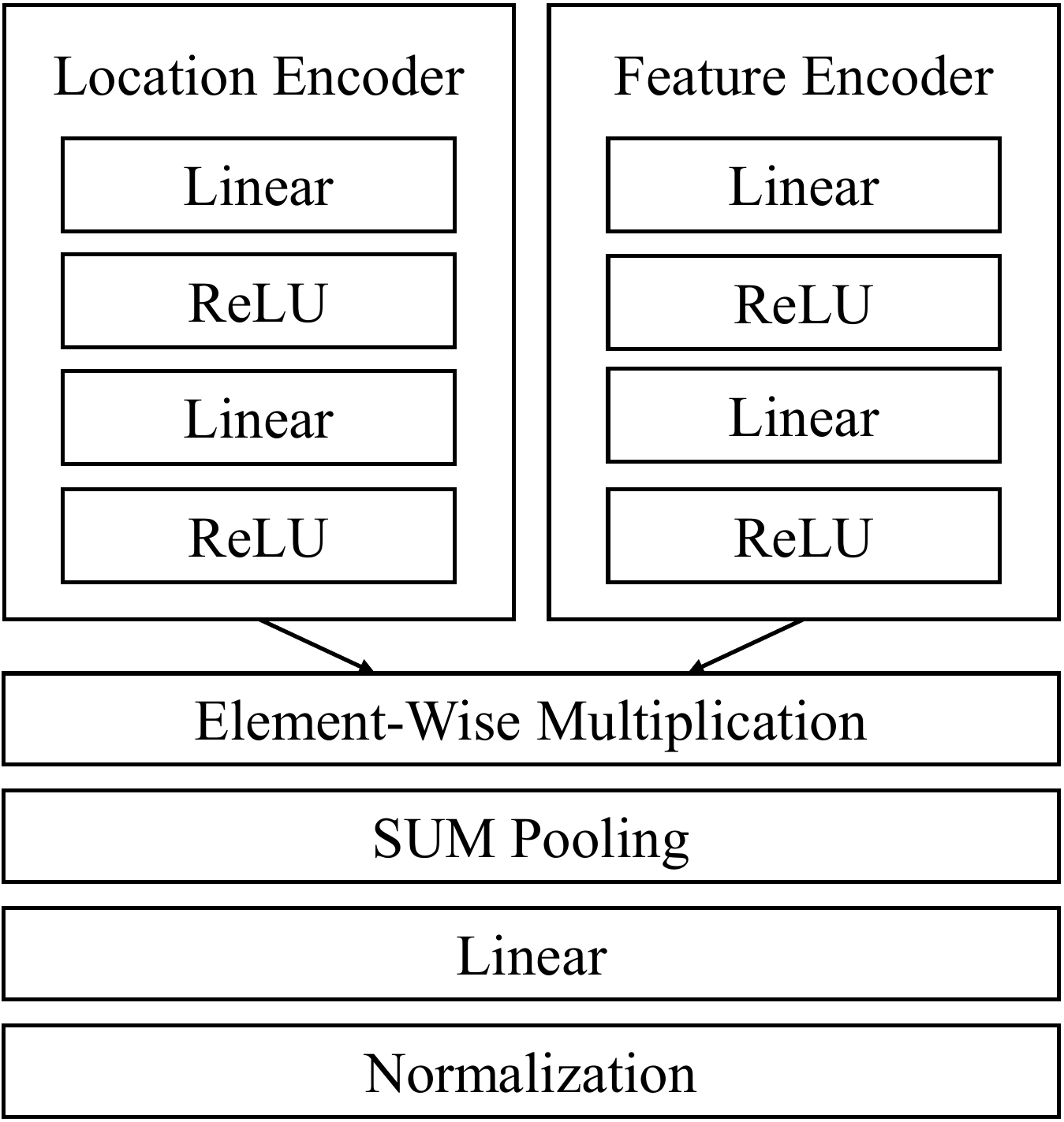}
    \caption{\fix{The structure of sparsity module, containing two parallel branches for location and content features, respectively.}}
    \label{fig:sparsity-module}
\end{figure}

Inspired by the (\rom{2}) property mentioned in \sref{sec:motivation}, we argue that a key-point feature should only be able to affect the object descriptor on its sparse locations, so that the key-points can be added or removed without significantly changing the object descriptor.
Therefore, to fully utilize the expressive power, key-points with different local patterns should be written into distinctive locations in the object descriptor.
Therefore, the sparse locations should be learned from the contents of the node embeddings. To achieve this, we define two branches for the sparsity layer in \eqref{eq:sparsity-layer} for the location and content features, respectively.
\begin{subequations}\label{eq:sparsity-layer}
    \begin{align}\label{eq:location-feature}
        ^{(l+1)}\mathbf{x}_i^L & = \operatorname{ReLU} (\mathbf{W}_L^{(l)} \cdot ^{(l)}\mathbf{x}_i^L),\\
        ^{(l+1)}\mathbf{x}_i^C & = \operatorname{ReLU} (\mathbf{W}_C^{(l)} \cdot ^{(l)}\mathbf{x}_i^C),
    \end{align}
\end{subequations}
where the superscript $(l), (l+1)$ indicates their layer number, $\mathbf{W}_L^{(l)},\mathbf{W}_C^{(l)}\in \mathbb{R}^{N_o \times N_n^{l}}$~($N_n^{l} < N_o)$  are the learnable location and content weights, respectively.
We set the input of \eqref{eq:sparsity-layer} as the same, \ie
$^{(1)}\mathbf{x}_i^L=^{(1)}\mathbf{x}_i^C$, which is the node embedding from the attention graph.
Specifically, each of the two branches has two independent layers. We present a more detailed structure for the proposed sparsity module in \fref{fig:sparsity-module}.

Intuitively, if the location feature $\mathbf{x}_i^L$ is a sparse vector that only a few locations are non-zeros, an object descriptor $\mathbf{D}_k$ in \eqref{eq:object-descriptor} will be insensitive to the change of key-points.
\begin{equation}\label{eq:object-descriptor}
    \mathbf{D}_k  = \phi({\mathbf{W}_o}\sum_{i=1}^{M} \mathbf{x}_i^L \odot \mathbf{x}_i^C + \mathbf{b}_o),
\end{equation}
where $\mathbf{W}_O$, $\mathbf{b}_O$ are learnable parameters and the operator $\odot$ is an element-wise multiplication. Note that the summation in \eqref{eq:object-descriptor} is a symmetric operator, which can ensure the object descriptor $\mathbf{D}_k$ to be invariant to the permutation of key-points.

It is intuitive that fewer number of non-zeros in the location feature $\mathbf{x}_i^L$ means that the object descriptor $\mathbf{D}_k$ will be less affected by the key-point $i$. We next show that how can we ensure the sparsity of the location feature $\mathbf{x}_i^L$.

\subsection{Loss Functions}

\subsubsection{Sparse Location Loss} \label{sec:sparsity-loss}

We define the sparse location loss $L_s$ as the $\ell_1$-norm of $\mathbf{x}_i^L$ to promote the sparsity:
\begin{equation}\label{eq:sparse-loss}
    L_s := \sum_{i=1}^{M} \left\|\phi({\mathbf{x}}_i^L)\right\|_1,
\end{equation}
where $\phi(\mathbf{x}) = \nicefrac{\mathbf{x}}{\left\|\mathbf{x}\right\|_2} $ is a $\ell_2$-normalization function, which is to prevent the location feature from being zero.
The $\ell_1$ loss for promoting vector sparsity has been widely used in sparse coding for dictionary learning \cite{lee2007efficient}.
An intuitive geometric explanation in 2-D space is that the points located on the intersection of axes and unit circle has the minimum $\ell_1$-norm, which means that some coordinates tend to be zero.

\begin{figure*}[!t]
    \centering
    \subfloat[PRC on OTB dataset.]
    {\includegraphics[width=0.32\linewidth]{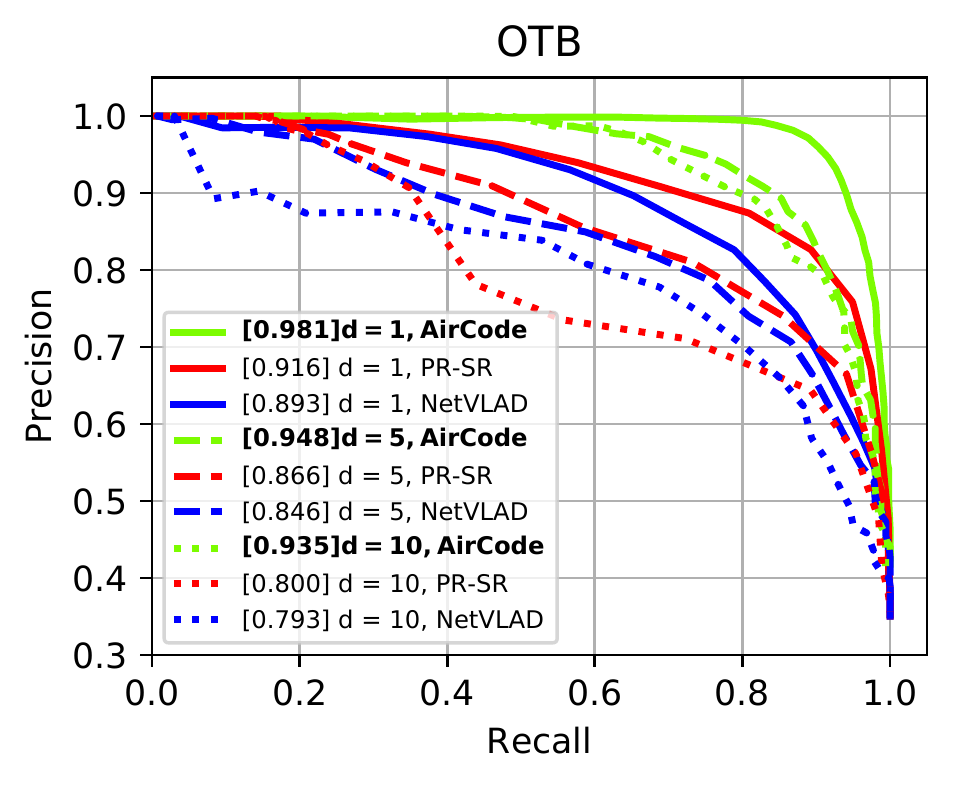}%
        \label{fig:otb}
    }
    \hfil
    \subfloat[PRC on VOT dataset.]
    {\includegraphics[width=0.32\linewidth]{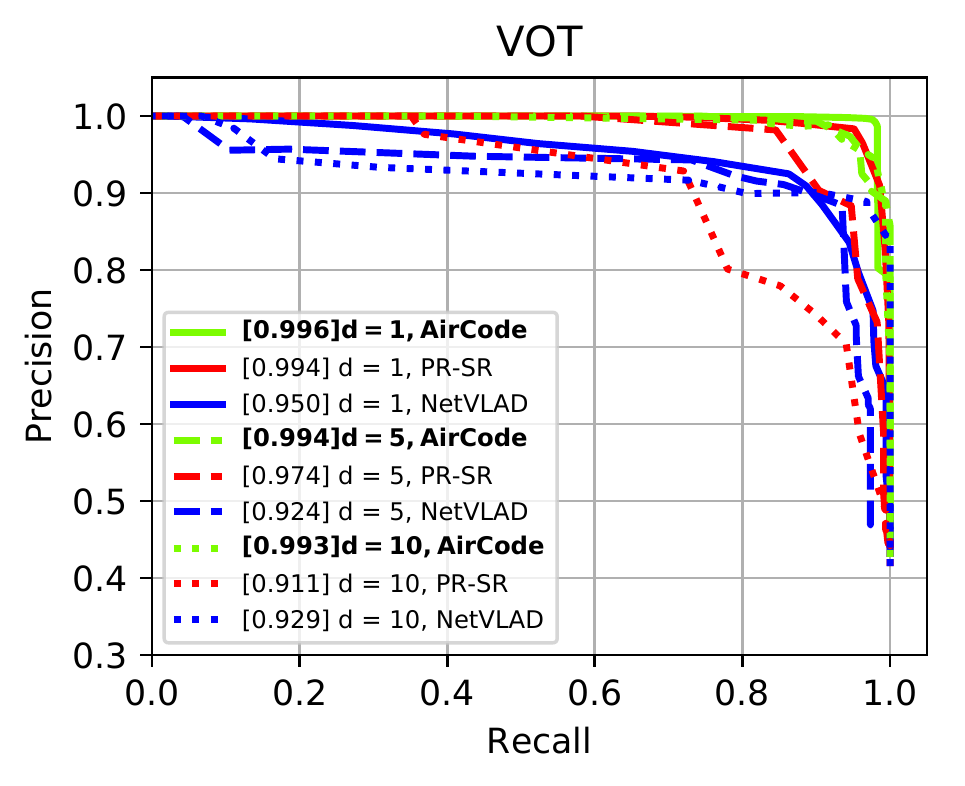}%
        \label{fig:vot}
    }
    \hfil
    \subfloat[PRC on KITTI Tracking dataset.]
    {\includegraphics[width=0.32\linewidth]{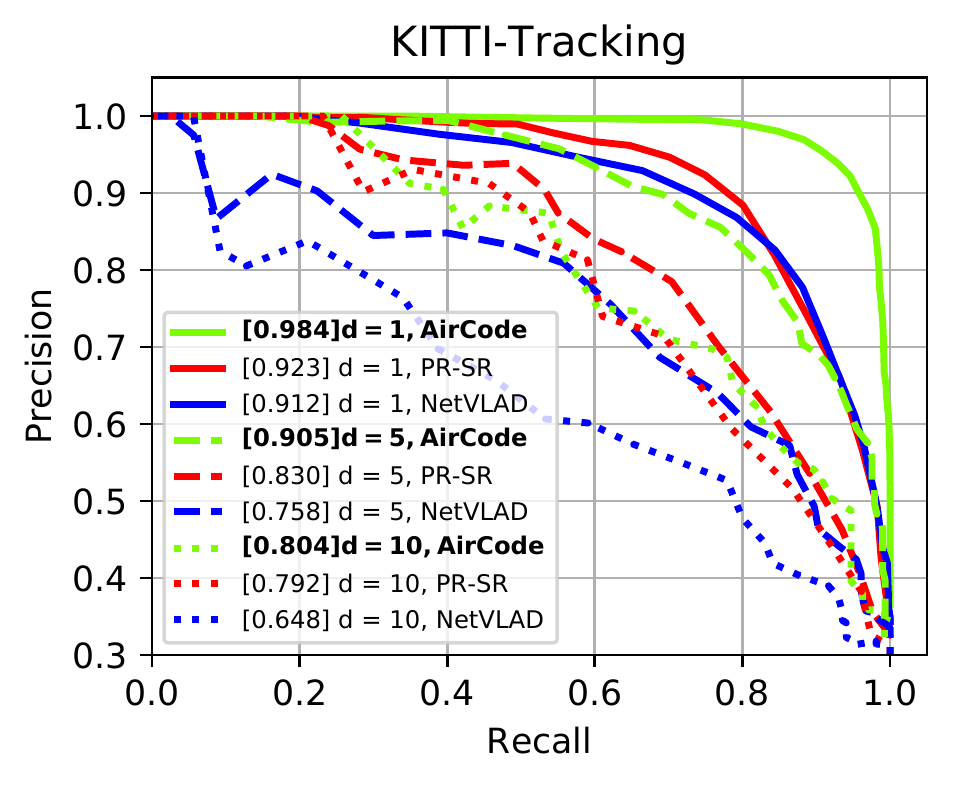}%
        \label{fig:kitti}
    }
    \caption{The precision-recall curves (PRC) on object matching. The area under curves are shown in the [brackets].}
    \label{fig:tracking}
\end{figure*}

\subsubsection{Dense Feature Loss} \label{sec:dense-loss}

The location sparse loss ensures that the key-point descriptors are encoded into sparse locations in the object descriptor. On the other hand, we also expect that distinctive key-point features are able to cover different locations, which can maximize the location utilization rate of the object descriptor, otherwise the location weights $\mathbf{W}_L$ \fix{may} simply learn to map all key-points to the same sparse locations, which is not expected.
Therefore, we define an object dense loss $L_d$ as the negative $\ell_1$-norm of the summed location features to improve the density of the object descriptor.
\begin{equation}\label{eq:dense-loss}
    L_d := \max \left(0, \delta- \phi\left( \left\|\sum^{M}_{i=1} \mathbf{x}_i^L \right\|_1\right)\right),
\end{equation}
where $\delta>0$ is a positive constant. A major difference of dense loss \eqref{eq:dense-loss} from sparse loss \eqref{eq:sparse-loss} is that the summation operator is inside of the $\ell_1$-norm, while the $\ell_2$-normalization function $\phi$ is outside of the $\ell_1$-norm.
Intuitively, by optimizing the two loss functions simultaneously, we are able to push the key-point location features to be sparse, while retaining the density of object descriptor, so that similar key-points will share similar locations, while distinctive key-points tend to cover different locations.
This means that the object descriptor is insensitive to the changes of key-points, resulting in the invariance to viewpoint changes and object deformation.

Note that the sparsity in this paper is different from the traditional sparse coding techniques \cite{lee2007efficient} and deep sparse coding network \cite{gwon2016deep} which are to learn a dictionary and minimize a reconstruction error, while ours is to learn sparse features without any approximation and dictionary.
It is also different from the model compression networks such as \cite{liu2020layerwise}, which are to learn sparse convolutional weights, as we have no sparse constraints on the learnable weights $\mathbf{W}_{L}$.

\subsubsection{Object Matching Loss}\label{sec:matching-loss}

The object matching loss $L_m$ maximizes the cosine similarity for positive object pairs and minimizes the cosine similarity for negative pairs.
\begin{equation}\label{eq:matching-loss}
    \begin{aligned}
         L_m :=& \sum_{\{p,q\}\in P^+} (1- S\left(\mathbf{D}_p,  \mathbf{D}_q\right))  \\
         +& \sum_{\{p,q\}\in P^-}\max (0,S(\mathbf{D}_p,  \mathbf{D}_q)-\zeta),
    \end{aligned}
\end{equation}
where $\zeta=0.2$ is a constant margin, $S$ is the cosine similarity, and $P^+$/$P^-$ are positive/negative object pairs, respectively.

\section{Experiments}

\subsection{Implementation Details and Baseline}

We take pre-trained backbone CNN and key-point detector from SuperPoint \cite{detone2018superpoint}, pre-trained segmentation header from Mask R-CNN \cite{he2017mask}, and \fix{train the graph and sparsity modules only on COCO dataset} \cite{lin2014microsoft}.
Random homographies are generated for data augmentation, including the perspective, translation, rotation, and scale transforms. We take a batch size of 16, a learning rate of $10^{-5}$, and the RMSprop \cite{hinton2012neural} optimizer for training AirCode.
In the experiments, we set $N_p= 256$, $N_m=16$, and $N_o=2048$.
We set the loss weights as 1.0, 0.5, 0.1, and 10 for negative matching, positive matching, sparse, and dense loss functions, respectively. The source code and pretrained models are available at \url{https://github.com/wang-chen/AirCode}.

We note that an attention graph is also used in SuperGlue \cite{sarlin2020superglue} to match feature points.
It broadcasts node features from two images for point matching, while object embedding in AirCode is self-representative thus the node features can only be propagated within the object.
Therefore, we don't take the point features from SuperGlue as our inputs, since we expect that an object descriptor doesn't use information from external images, so that one query descriptor can be matched to objects in database without exchanging information. 

We perform extensive comparison with the state-of-the-art methods including NetVLAD \cite{arandjelovic2016netvlad} and PR-SR \cite{wang2020online}.
NetVLAD is an image matching method and has been adopted by many place recognition algorithms, e.g., \cite{sarlin2019coarse,liu2019lpd}, which verifies its generalization ability. In the experiments, we use it as a feature aggregation method.
PR-SR applied a saliency detection technique to segment objects, which sometimes is not able to detect all objects.
For fair comparison, we provide the ground truth mask of object segmentation for PR-SR and only compare with its object matching performance.

\subsection{Object Matching}\label{sec:matching}

We first present the performance of the proposed method on object matching. Three datasets will be tested extensively, including OTB-2013 \cite{wu2013online}, VOT-2016 \cite{VOT_TPAMI}, and KITTI Tracking \cite{luiten2021hota}.
Note that we present performance on tracking datasets but don't compare with tracking methods such as \cite{wang2018kernel}, as they mainly use the information from consecutive frames, while we extract descriptors for matching with any frames.
The performance is reported without finetuning on these datasets.

\begin{figure*}[t]
    \centering
    \includegraphics[width=1\linewidth]{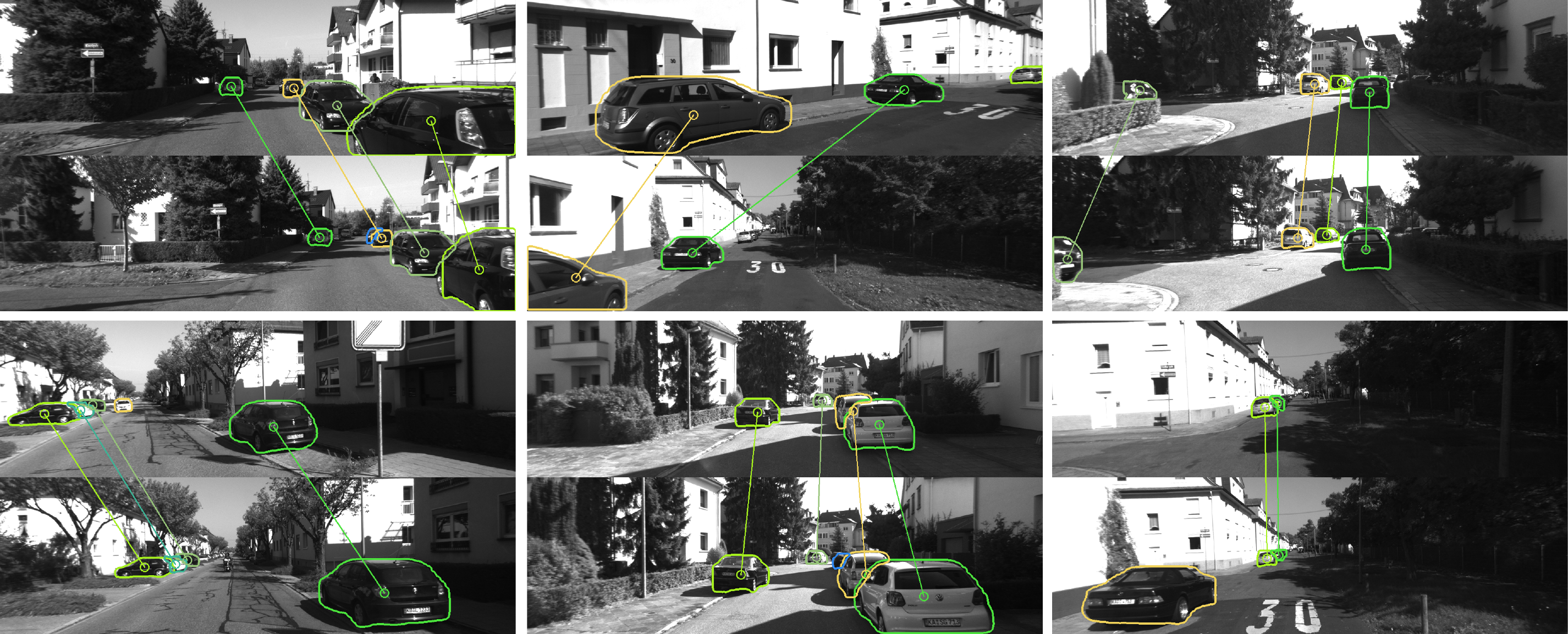}
    \caption{The examples of car matching in the KITTI Odometry dataset. Although all the cars look similar, AirCode is still able to correctly identify the same cars from different viewpoints.}
    \label{fig:kitti-relocalization}
\end{figure*}

\begin{table}[t]
    \caption{Object Matching on the OTB dataset.}
    \label{tab:otb}
    \centering
    \begin{tabular}{C{0.1\linewidth}C{0.15\linewidth}C{0.13\linewidth}C{0.13\linewidth}C{0.13\linewidth}}
        \toprule
        &  Method & Precision & Recall & F1  \\ \midrule
        \multirow{3}{*}{$d=1$} & NetVLAD & 78.5 &	83.0 & 80.7  \\
        & PR-SR & 82.7 & 89.3 & 85.9 \\
        & AirCode & \textbf{93.2} & \textbf{92.6} & \textbf{92.9} \\ \midrule
        \multirow{3}{*}{$d=3$} & NetVLAD &72.7	& 79.7 &	76.0 \\
        & PR-SR & 81.7 & 79.1 & 80.4 \\
        &AirCode & \textbf{88.3} &	\textbf{86.4} &	\textbf{87.3} \\ \midrule
        \multirow{3}{*}{$d=5$} &  NetVLAD& 74.0	& 81.0 & 77.3   \\
        & PR-SR & 81.0 & 73.2 & 76.9 \\
        & AirCode &  \textbf{87.6} &\textbf{ 86.2} & \textbf{86.9}  \\ \midrule
        \multirow{3}{*}{$d=10$} & NetVLAD &  70.2 &	80.7 & 75.1 \\
        & PR-SR & 71.1 & 72.2 & 71.6 \\
        & AirCode &  \textbf{85.6} & \textbf{84.7} & \textbf{85.1}  \\
        \bottomrule
    \end{tabular}
\end{table}

\subsubsection{OTB Dataset}

We select all sequences of OTB in which objects have labels in the COCO dataset and contain 5 or more key-points.
To test the robustness, we report the matching performance for frame pairs that have $d-1$ ($d>1$) frames in between. Intuitively, matching is more difficult for larger $d$ and object tracking is the case of $d=1$.
We take an object pair as a match if their cosine similarity is larger than a threshold $\delta$.
The overall performance of precision, recall, and F1-score is presented in \tref{tab:otb}. It can be seen that AirCode outperforms NetVLAD with a large margin of 5\%-15\% and outperforms PR-SR by a margin of 3\%-10\%.

Intuitively, the precision-recall curves (PRC) can be obtained for each sequence by changing the threshold $\delta$.
\fref{fig:otb} shows the PRC for all datasets, in which the areas under precision-recall curves (AU-PRC) are reported in the brackets. AirCode achieves a much higher performance for all cases, i.e., an average higher performance of 8.8\%, 10.2\%, and 14.2\% than NetVLAD and 6.5\%, 8.2\%, and 13.5\% than PR-SR for $d=1$, $5$, and $10$, respectively.
We notice that the performance of AirCode decreases much slower with increasing $d$ than other methods, which verifies its robustness.

\begin{table}[t]
    \caption{Object Matching on the VOT dataset.}
    \label{tab:vot}
    \centering
    \begin{tabular}{C{0.1\linewidth}C{0.15\linewidth}C{0.13\linewidth}C{0.13\linewidth}C{0.13\linewidth}}
        \toprule
        &  Method & Precision & Recall & F1  \\ \midrule
        \multirow{3}{*}{$d=1$} & NetVLAD & 90.9	& 88.7 & 89.8  \\
        & PR-SR & 98.3 & 95.1 & 96.7 \\
        & AirCode & \textbf{98.6} & \textbf{98.3} & \textbf{98.4} \\ \midrule
        \multirow{3}{*}{$d=3$} & NetVLAD &90.0 & 92.1 & 91.0\\
        & PR-SR & 97.2 & 94.2 & 95.7 \\
        &AirCode & \textbf{97.4} & \textbf{98.2} & \textbf{97.8} \\ \midrule
        \multirow{3}{*}{$d=5$} &  NetVLAD& 89.5 & 90.4 & 89.9 \\
        & PR-SR & 90.3 & 90.4 & 90.3 \\
        & AirCode &  \textbf{95.8} & \textbf{95.9} & \textbf{95.8}  \\ \midrule
        \multirow{3}{*}{$d=10$} & NetVLAD &  90.0 &	90.5 & 90.2 \\
        & PR-SR & 77.9 & 85.1 & 81.3 \\
        & AirCode &  \textbf{95.6} & \textbf{95.3} & \textbf{95.4}  \\
        \bottomrule
    \end{tabular}
\end{table}

\subsubsection{VOT Dataset}

The VOT dataset mainly contains blurred and non-rigid objects such as humans.
The overall performance is presented in \tref{tab:vot}, in which we achieve a higher performance of 1-5\% than NetVLAD and comparable performance with PR-SR.
The PRC performance is reported in \fref{fig:vot}, in which we achieve a higher performance of 4.6\%, 7.0\%, and 6.4\%, for $d=1$, $5$, and $10$ than NetVLAD and 0.2\%, 2.0\%, and 8.2\% than PR-SR for $d=1$, $5$, and $10$, respectively.s.

\subsubsection{KITTI Tracking}

The KITTI Tracking is a multi-object tracking dataset \cite{luiten2021hota}, thus we calculate the similarity of all object pairs in each frame pair.
The overall performance on precision, recall, and F1-score are reported in \tref{tab:kitti} in which AirCode still achieves the best performance. Their PRC are shown in \fref{fig:kitti}, in which we achieve a higher performance of 7.2\%, 14.7\%, and 15.6\% than NetVLAD and 6.1\%, 7.5\%, and 1.2\% than PR-SR for $d=1$, $5$, and $10$, respectively.

\begin{table}[t]
    \caption{Object Matching on the KITTI Tracking dataset.}
    \label{tab:kitti}
    \centering
    \begin{tabular}{C{0.1\linewidth}C{0.15\linewidth}C{0.13\linewidth}C{0.13\linewidth}C{0.13\linewidth}}
    \toprule
    &  Method & Precision & Recall & F1  \\ \midrule
    \multirow{3}{*}{$d=1$} & NetVLAD & 82.6 & 84.4 & 83.5 \\
    & PR-SR & 82.0 & 84.3 & 83.1 \\
    & AirCode & \textbf{93.8} & \textbf{93.0} & \textbf{93.4} \\ \midrule
    \multirow{3}{*}{$d=3$} & NetVLAD &72.7 & 74.1& 73.4 \\
    & PR-SR & 79.4 & 74.5 & 76.9 \\
    &AirCode & \textbf{89.1} & \textbf{85.8} & \textbf{87.4} \\ \midrule
    \multirow{3}{*}{$d=5$} &  NetVLAD & 68.6 & 68.8 & 68.7  \\
    & PR-SR & 78.5 & 70.4 & 74.2 \\
    & AirCode &  \textbf{82.3} & \textbf{80.6} & \textbf{81.4}  \\ \midrule
    \multirow{3}{*}{$d=10$} & NetVLAD &  57.4 & 65.3 & 61.1 \\
    & PR-SR & \textbf{71.4} & 69.3 & 70.3 \\
    & AirCode &  71.0 & \textbf{69.9} & \textbf{70.4} \\
    \bottomrule
\end{tabular}
\end{table}

\subsection{Semantic Relocalization}\label{sec:relocalization}

To further verify its robustness, we apply the proposed method to semantic relocalization, which will be evaluated on the KITTI Odometry dataset \cite{geiger2012we}.  We select the sequences with significant loop closures, i.e.,  '00', '05', and '06'.
\fix{To calculate the similarity of two images, we accumulate the similarities of all the object pairs whose similarities are larger than a threshold.} 
To test the robustness of our method, we mainly re-identify cars for relocalization, which is much more difficult than matching distinct objects. Although the overall performance of relocalization will be slightly harmed, such a setting is able to verify the robustness of the object descriptor, as the cars on the road in KITTI datasets are very similar thus the performance on car matching is more convincing.

\begin{table}[t]
    \caption{Semantic Relocalization on KITTI Odometry.}
    \label{tab:relocalization}
    \centering
    \begin{tabular}{C{0.1\linewidth}C{0.15\linewidth}C{0.13\linewidth}C{0.13\linewidth}C{0.13\linewidth}}
        \toprule
        Sequence &  Method & Precision & Recall & F1  \\ \midrule
        \multirow{2}{*}{00} & NetVLAD & 21.1 & 41.3 & 27.9 \\
        & AirCode & \textbf{95.7} & \textbf{75.4} & \textbf{84.3} \\ \midrule
        \multirow{2}{*}{05} & NetVLAD & 19.0 & 25.3 & 21.7 \\
        &AirCode & \textbf{97.2} & \textbf{40.1} & \textbf{56.8} \\ \midrule
        \multirow{2}{*}{06} &  NetVLAD & 39.0 & 27.4 & 32.2 \\
        & AirCode &  \textbf{95.4} & \textbf{47.7} & \textbf{63.6} \\
        \bottomrule
    \end{tabular}
\end{table}

\begin{figure}[th]
    \centering
    \includegraphics[width=0.9\linewidth]{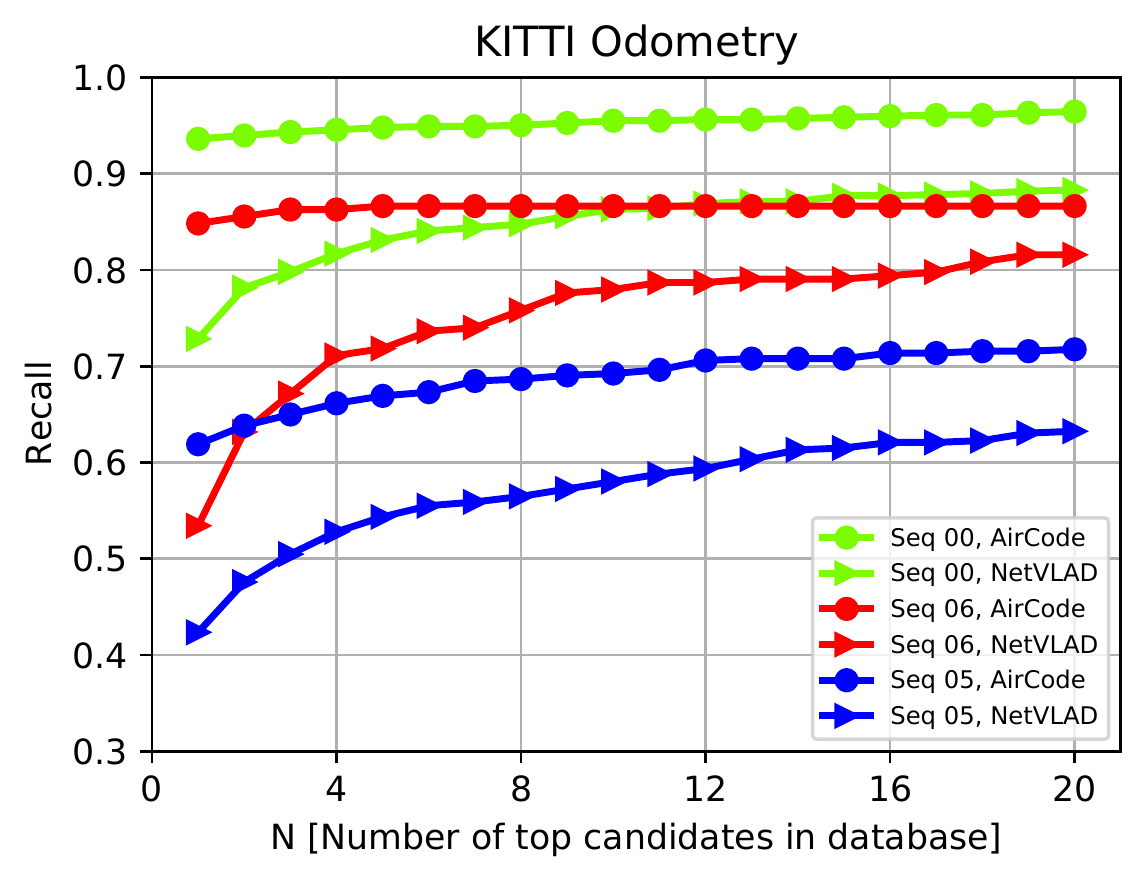}
    \caption{The performance of semantic re-localization.}
    \label{fig:kitti-odometry}
\end{figure}

The overall performance is listed in \tref{tab:relocalization}, in which we can see that we achieve more reliable performance than NetVLAD for all sequences.
Note that we remove assistant techniques such as geometric verification \cite{wang2020online} to ensure that the improvements come from the proposed object descriptor. We also present the recall curves in \fref{fig:kitti-odometry}, in which we accept the top $N$ pairs as a re-localization. It can be seen that, $\forall N$, $N\in[1,20]$, we are able to achieve better performance with a large margin than NetVLAD for all sequences, which verifies the effectiveness of AirCode. PR-SR is excluded from this task, since it computes correlation for all object pairs, which is quite computationally heavy.

\subsection{Live Demo \& Limitation}

\begin{figure}[t]
    \centering
    % \vspace{-5pt}
    \subfloat[Change of Viewpoint.]
    {\includegraphics[width=0.485\linewidth]{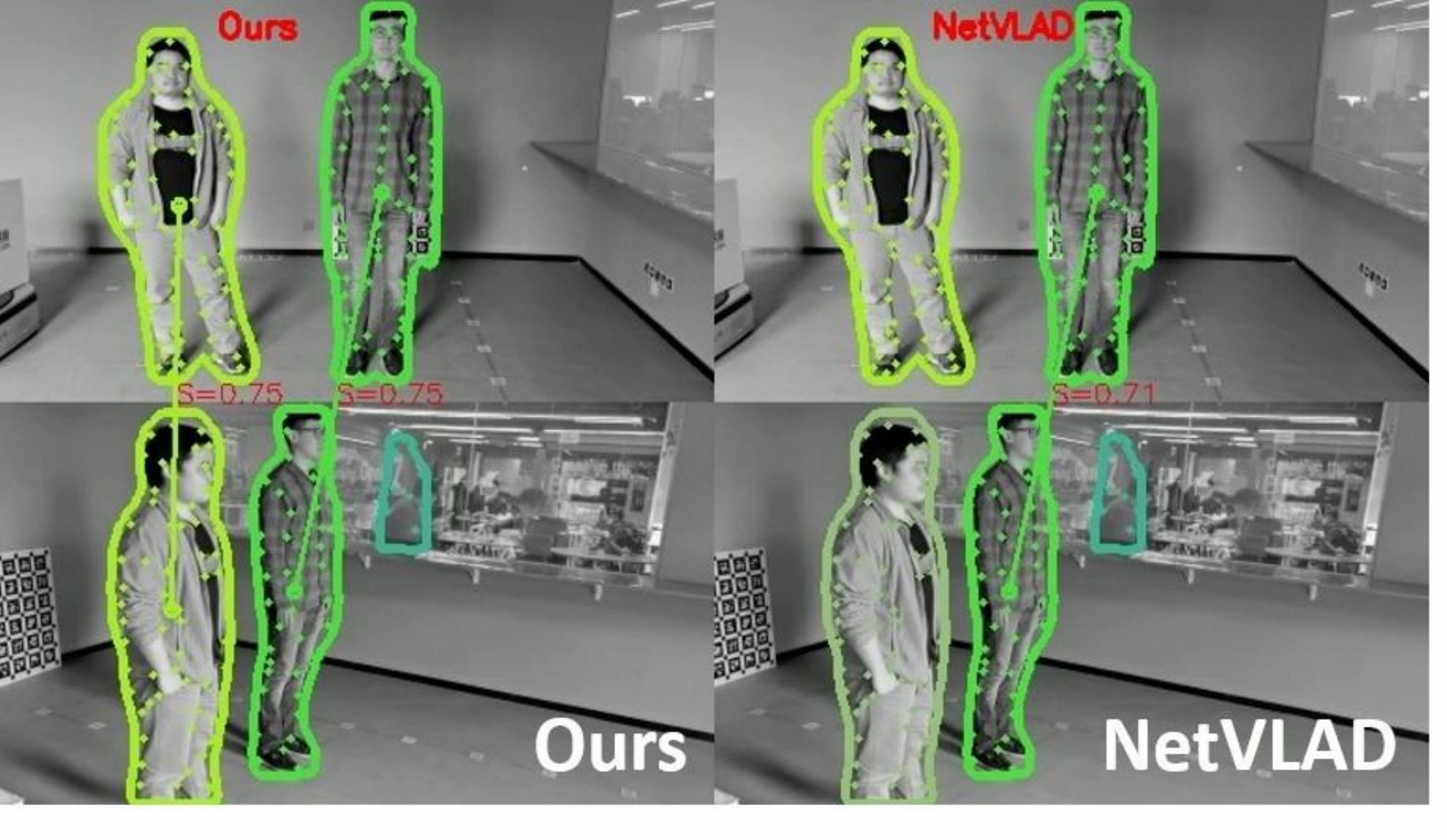}%
        \label{fig:viewpoint}
    }
    \hfil
    \subfloat[Object Occlusion.]
    {\includegraphics[width=0.485\linewidth]{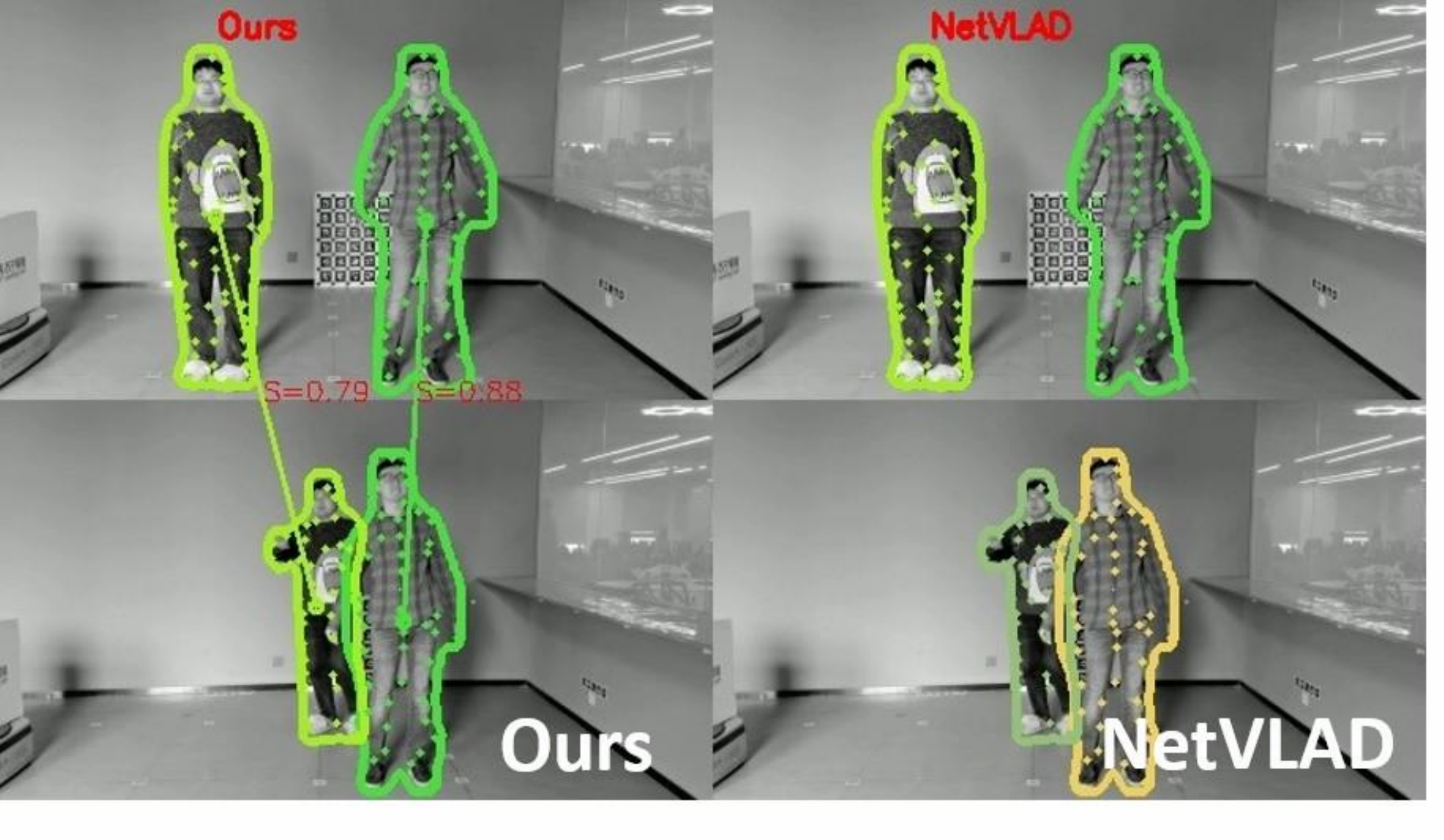}%
        \label{fig:occupied}
    }
    
    \subfloat[Object Deformation \rom{1}.]
    {\includegraphics[width=0.485\linewidth]{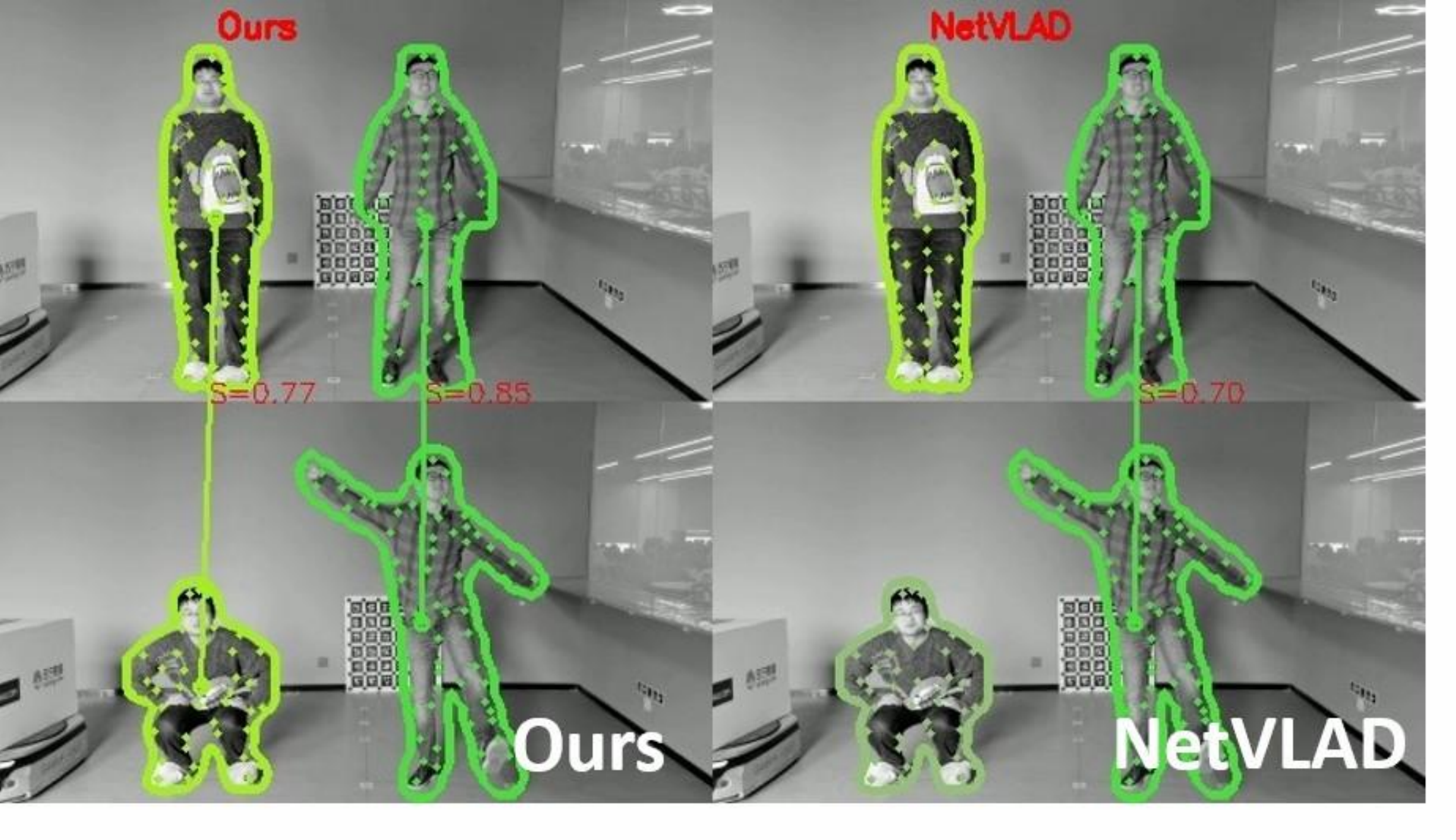}%
        \label{fig:deformation1}
    }
    \subfloat[Object Deformation \rom{2}.]
    {\includegraphics[width=0.485\linewidth]{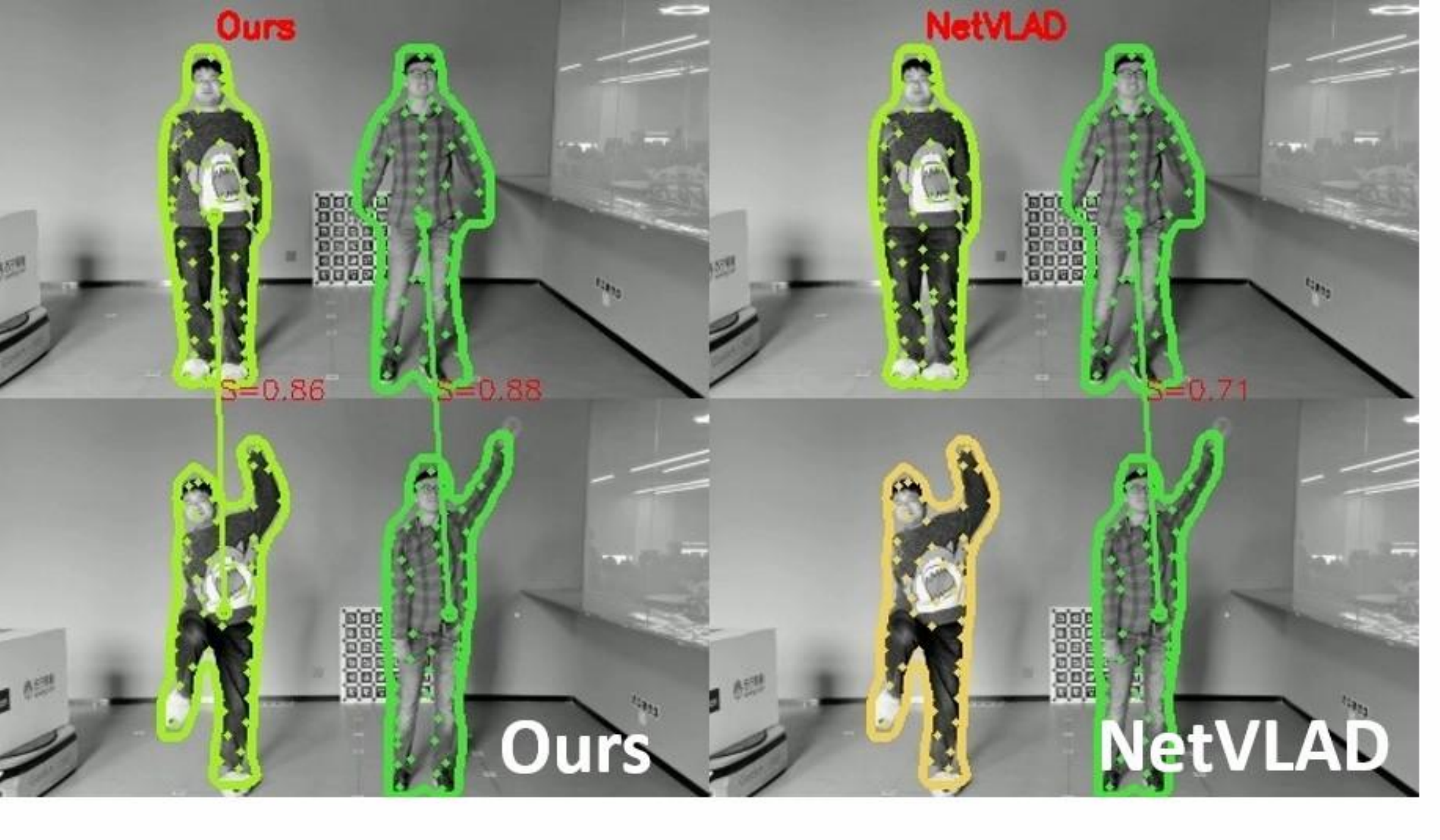}%
        \label{fig:deformation2}
    }
    \caption{The live object matching demo.}
    \label{fig:livedemo}
\end{figure}

In this section, we present a live demo to demonstrate the generalization ability and robustness of the proposed method.
In this demo, we still use the pre-trained model mentioned in \sref{sec:matching} without any fine-tuning.
It can be seen that, compared to NetVLAD, AirCode is quite robust to the change of viewpoint in \fref{fig:viewpoint}, object occlusion in \fref{fig:occupied}, and object deformation in \fref{fig:deformation1} and \ref{fig:deformation2}.
For better visualization, we strongly suggest the readers watch the video attached to this paper at \url{https://youtu.be/ZhW4Qk1tLNQ}.

\subsection{Sparsity Analysis}

\begin{figure}[t]
    \centering
    \subfloat[Key-point Sparsity.]
    {\includegraphics[width=0.48\linewidth]{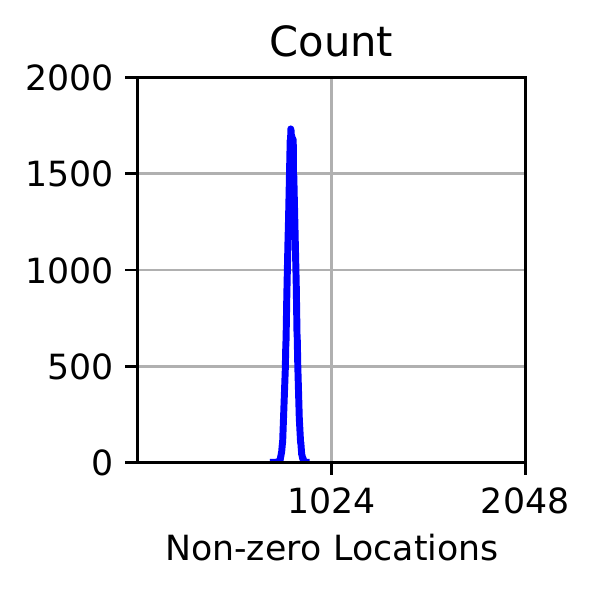}%
        \label{fig:point}
    }
    \hfill
    \subfloat[Average Object Denseness]
    {\includegraphics[width=0.48\linewidth]{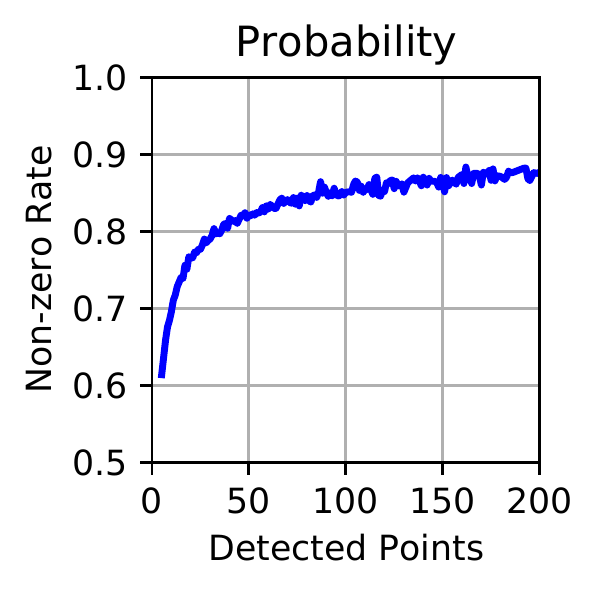}%
        \label{fig:object}
    }
    \caption{The statistics of feature sparsity.}
    \label{fig:sparsity}
\end{figure}

We provide the sparsity analysis for the proposed method based on COCO dataset \cite{lin2014microsoft}.
The distribution of non-zero locations of the key-point features is shown in \fref{fig:point}, in which we can see that only 700-800 out of 2048 locations are non-zeros.
This verifies the objective of sparse loss in \eqref{eq:sparse-loss}.
The statistics of non-zero locations of the object descriptor are presented in \fref{fig:object}.
As expected, the non-zero locations increase with the number of key-points detected in the object, which also verifies the objective of dense loss \eqref{eq:dense-loss}.

\fix{
Note that we expect to group similar key-points into the same locations as in \eqref{eq:location-feature}, so that the location of a single object descriptor will not be quickly used up by simply increasing the number of key-points, since an object typically cannot cover all different points.
This can be further verified by calculating the location usage over different object descriptors.
To this end, we randomly select $N$ images and calculate how often each location of the object descriptor is used. 
We list the usage rate for $N=1$, $10$, $100$, $1000$, and $10000$ in \tref{tab:occupied_rate}.
It can be seen that the location of object descriptor is nearly fully used by increasing the number of objects, which further verifies the effectiveness of the dense feature loss.
}

\begin{table}[h]
 \caption{\fix{The Location Usage of Object Descriptors.}}
 \label{tab:occupied_rate}
 \centering
 \begin{tabular}{cccccc}
     \toprule
     N & 1 & 10  &  100 &  1000 & 10000 \\
     \midrule
     Usage Rate & 0.868 & 0.927 & 0.966 & 0.979 & 0.984 \\
     \bottomrule
 \end{tabular}
\end{table}

\subsection{Efficiency}

This section presents the runtime of the proposed modules, which are listed in \tref{tab:running}. It can be seen that the sparsity layer only takes about 1 \milli\second, which is much lower than the graph attention network used in this module. The overall running time is about 9.3 \milli\second, which satisfies the real-time requirements of most robotic applications.

 \begin{table}[!ht]
     \caption{Runtime Analysis.}
     \label{tab:running}
     \centering
     \begin{tabular}{ccccc}
         \toprule
         Module & Node Encoding & Graph  &  Sparsity &  Overall \\
         \midrule
         Time & 0.847 \milli\second & 7.302 \milli\second & 1.131 \milli\second & 9.280 \milli\second \\
         \bottomrule
     \end{tabular}
 \end{table}

\fix{
\section{Ablation Study}

\subsection{Sparsity Module}

To analyze the effectiveness of the sparsity module, we compare with a model by replacing it with a fully connected layer to output a descriptor with the same dimension as the original one.
Similar to \sref{sec:matching} and \sref{sec:relocalization}, the performance on object matching ($d=10$) and semantic relocalization are evaluated.
The results are listed in the 2nd and 4th row of \tref{tab:ablation_object_matching} and \tref{tab:ablation_relocalization}, respectively, where ``w/o sparsity'' denotes the model without sparsity module.
It can be seen that the sparsity layer brings a huge overall performance gain, especially in the task of semantic relocalization.
It is also observed that, AirCode w/o sparsity performs a little bit better in object matching on OTB dataset. This is because OTB is relatively an easier dataset, so the model with a fully connected layer can also work well.

\begin{table}[t]
    \caption{\fix{The Ablation Study on Object Matching.}}
    \label{tab:ablation_object_matching}
    \centering
    \begin{tabular}{C{0.25\linewidth}C{0.12\linewidth}C{0.12\linewidth}C{0.22\linewidth}}
        \toprule
        Dataset & OTB & VOT & KITTI Tracking\\
        \midrule
        AirCode$^{\text{w/o sparsity}}$ & \textbf{94.0} & 98.6 & 79.7 \\ 
        AirCode$^{\text{w/o loss}}$ &91.7&98.8 & 75.1  \\  
        AirCode & \underline{93.5} & \textbf{99.3} & \textbf{80.4} \\ 
        \bottomrule
    \end{tabular}
\end{table}

\subsection{Loss Functions}

We next explore that effects of the proposed sparse and dense loss functions in \eqref{eq:sparse-loss} and \eqref{eq:dense-loss}. Note that the two loss functions should be used simultaneously as they need to balance each other, so we remove both of them in this experiment. The 3rd and 4th row of \tref{tab:ablation_object_matching} and \tref{tab:ablation_relocalization} show that removing the loss functions leads to a significant drop of overall performance in both tasks. In the semantic relocalization experiment, AirCode also achieves an average higher F1-score of 4\% than AirCode w/o loss, which verifies the necessity of the proposed sparse and dense loss functions. We notice that AirCode w/o loss has a higher recall. The reason is that every point feature in AirCode is sparser, so the final object encoding is more discriminative, which leads to a higher precision but lower recall with the same threshold.

\begin{table}[t]
    \caption{\fix{The Ablation Study on Semantic Relocalization.}}
    \label{tab:ablation_relocalization}
    \centering
    \begin{tabular}{C{0.25\linewidth}C{0.15\linewidth}C{0.15\linewidth}C{0.15\linewidth}}
        \toprule
         Metric & Precision & Recall & F1  \\ \midrule
        AirCode$^{\text{w/o sparsity}}$ & 90.3 & 37.1 & 52.6 \\
        AirCode$^{\text{w/o loss}}$ & 76.4 & \textbf{57.2} & 65.4 \\
        AirCode &  \textbf{96.0} & \underline{54.4} & \textbf{69.4} \\
        \bottomrule
    \end{tabular}
\end{table}

\section{Limitation}

Despite the promising prospect and outstanding performance, AirCode still has several limitations.
First, its robustness depends on the performance of pre-trained segmentation header and point detector, thus its performance may not be stable when objects are segmented or the feature points are detected incorrectly.
Second, the object categories also depend on the object detector, thus it is difficult for AirCode to match novel objects.
Third, with robot moving, we often observe an object from multiple viewpoints, while AirCode is designed for object encoding on a single image.
Therefore, it can be further improved by introducing temporally ``evolving'' object representation, which will be taken as our future work.

}

\section{Conclusion}

In this work, we present a novel object encoding method, which can be used in many robotic tasks such as autonomous exploration and semantic relocalization.
% It is a plug-and-play module and can be easily adopted to any framework.
To be robust to the number of key-points detected, we propose a sparse encoding method to ensure that each key-point can only affect a small part of the object descriptors, making the global descriptors robust in severe conditions such as viewpoint changes and object deformation.
In the experiments, we show that it achieves superior performance in object matching and provides reliable semantic relocalization.
We also show a live demo to demonstrate its robustness. We expect that this method will play an important role in robotic applications.

%                                  % on the last page of the document manually. It shortens
%                                  % the textheight of the last page by a suitable amount.
%                                  % This command does not take effect until the next page
%                                  % so it should come on the page before the last. Make
%                                  % sure that you do not shorten the textheight too much.

%%%%%%%%%%%%%%%%%%%%%%%%%%%%%%%%%%%%%%%%%%%%%%%%%%%%%%%%%%%%%%%%%%%%%%%%%%%%%%%%

%%%%%%%%%%%%%%%%%%%%%%%%%%%%%%%%%%%%%%%%%%%%%%%%%%%%%%%%%%%%%%%%%%%%%%%%%%%%%%%%



% \section*{ACKNOWLEDGMENT}

% The preferred spelling of the word ÒacknowledgmentÓ in America is without an ÒeÓ after the ÒgÓ. Avoid the stilted expression, ÒOne of us (R. B. G.) thanks . . .Ó  Instead, try ÒR. B. G. thanksÓ. Put sponsor acknowledgments in the unnumbered footnote on the first page.

{
    \balance
    \bibliographystyle{IEEEtran}
    \bibliography{papers}
}

\end{document}